\pdfoutput=1 
\documentclass{article}
\usepackage{iclr2026_conference,times}
\newif\ifsubmission
\ifsubmission\else\iclrfinalcopy\fi
\providecommand{\nolinenumbers}{}
\providecommand{\linenumbers}{}

\usepackage[T1]{fontenc}
\usepackage[utf8]{inputenc}
\usepackage{microtype}
\usepackage{inconsolata}
\usepackage{hyperref}
\usepackage{url}
\usepackage{booktabs}
\usepackage{amsfonts}
\usepackage{amsmath}
\usepackage{amssymb}
\usepackage{graphicx}
\graphicspath{{figures/}{./}}
\usepackage{xcolor}
\usepackage{xspace}
\usepackage{fontawesome5}
\usepackage[most]{tcolorbox}
\definecolor{boxframe}{RGB}{170,170,170}
\definecolor{boxtitlebg}{RGB}{208,208,208}
\definecolor{tocblue}{RGB}{40,70,160}
\tcbset{graybox/.style={
  breakable, enhanced, colback=gray!3, colframe=boxframe,
  colbacktitle=boxtitlebg, coltitle=black, fonttitle=\bfseries\small,
  arc=0.6mm, boxrule=0.4pt, left=2mm, right=2mm, top=1mm, bottom=1mm,
  before skip=3pt, after skip=3pt}}
\newtcolorbox{promptbox}[1][]{graybox, fontupper=\footnotesize\ttfamily, title={#1}}

\title{They Infer What You Meant:\\ Models Represent Communicative Intent\\ More Reliably Than They Act On It}

\author{%
  Alex Kwon\\
  Independent Researcher\\
  \texttt{ask@collapseindex.org}
}

\begin{document}
\maketitle
\lhead{}  

\begin{abstract}
When a person shares something with a language model, the model often answers the \emph{surface} of the
message rather than what the sender was \emph{doing} by sending it: share a finished project and it critiques
the code; share a raw late-night line and it runs a wellness check. We treat the sender's communicative intent,
the Gricean \emph{what-was-meant}, as a first-class object of interpretability study, and show the failure is
one of readout on top of a robust representation. A linear probe decodes the sender's intent, whether they
want a thing \emph{recognized} or \emph{evaluated}, from a model's default-pass hidden states, cleanly and
surface-independently, across six models and four families and in the base checkpoints. The representation
generalizes further, to intent that is only pragmatically \emph{inferred}, and to a second, genuinely different
and lexically clean intent (support versus help). The behavioral half of the story, and every causal test, is
established on the recognize/evaluate contrast, where what varies is whether the default output acts on the
intent. The readout \emph{lags} the representation in depth within a model (the intent is decodable several
layers before it drives the output); across models, which ones act on it by default is model-specific, an
observed stratification (three of six show the failure) that we do not read as a scaling law. Where the gap is
open, a direction closely tied to the representation, the discriminative direction at a searched-for layer, is a
causal handle: steering it recovers the intended behavior, as well as an explicit instruction does and with no
prompt at all. This direction is near-orthogonal to the feedback-offering axis, so it routes a represented intent
rather than a generic feedback knob, though at the recovery dose the routed intent can override an explicit
request. We support each link with controls against the obvious deflations and report the nulls as plainly as the
confirmations.
\end{abstract}

\begin{figure}[t]
\centering
\includegraphics[width=\textwidth]{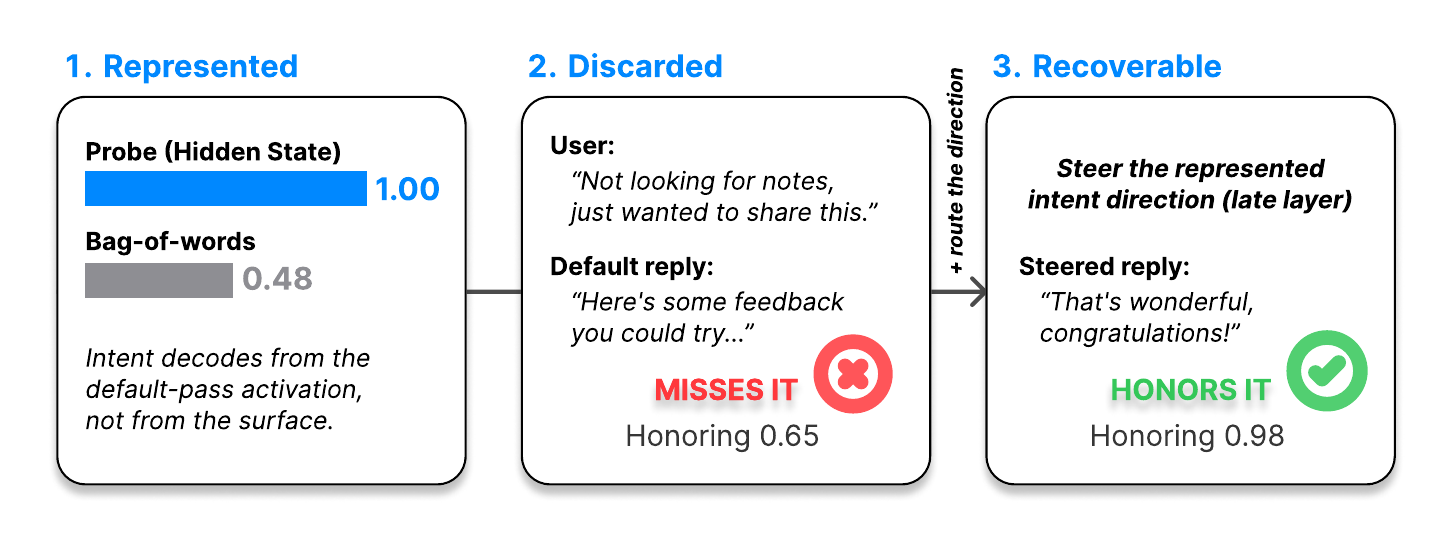}
\caption{\textbf{The represent-then-lagging-readout chain, exemplified on Qwen-3B (a model where the readout
gap is open).} \textbf{(1)}~A linear probe decodes the intent (recognize vs evaluate) from the default-pass
hidden state at $1.00$, while bag-of-words on held-out phrasings is at chance ($0.48$). \textbf{(2)}~The
default reply nonetheless offers unsolicited feedback, honoring a recognize-intent only about $0.65$ of the
time. \textbf{(3)}~Steering the residual stream along the discriminative intent direction at a late layer recovers
honoring to $0.98$, unsolicited feedback collapsing and coherence preserved. The representation is universal;
whether the readout acts on it is model-specific (Section~\ref{sec:scale}). The steered direction is
near-orthogonal to the feedback-behavior axis, routing a represented intent rather than a generic feedback knob
(Section~\ref{sec:spec}).}
\label{fig:teaser}
\end{figure}

\section{Introduction}
A recurring failure in deployed language models is that they respond to the literal content of a message and
miss its communicative intent, what the sender was trying to accomplish by sending it. The failure is hard to
see because the responses are fluent and often look caring: a model handed a person's first finished creative
project will, by default, offer improvements; a model handed a raw expression of exhaustion will, by default,
assess risk. Both responses pass a surface rubric, and both miss the person. The framing is Gricean \citep{grice1975logic}: a cooperative reader answers what
was \emph{meant}, not only what was said. Pragmatic competence of this kind has been \emph{benchmarked}
behaviorally \citep{ruis2023goldilocks}; we ask instead where the intent lives inside the model. A related, better-studied failure of instruction-tuned assistants
\citep{ouyang2022instructgpt} is sycophancy, deferring to a user's stated belief over the truth
\citep{sharma2023sycophancy}; ours is complementary, the default readout overrides the sender's \emph{goal},
not their stated belief.

Our finding is that the intent is robustly \emph{represented} and that the failure is one of \emph{readout}. A
linear probe reads the sender's intent out of the default-pass hidden state cleanly (Figure~\ref{fig:teaser}),
and it does so even when the intent is never stated and must be \emph{inferred} from context, so the model
performs the pragmatic inference internally. What varies is whether the readout acts on it, and it varies
systematically. The readout \emph{lags} the representation in depth: within a model, the intent is decodable
several layers before the layer at which it drives the output. Across models the pattern is a
\emph{stratification} rather than a lag, the same intent is represented everywhere but whether the default
readout acts on it is model-specific (the failure appears in three of six), which may reflect differing
feedback-offering priors as much as differing capability. This behavioral discard is not our headline claim but
our \emph{lens}: the regime where the represented-but-unused intent is visible in behavior and, crucially,
causally manipulable, steering a direction closely tied to it there recovers the honored behavior as well as an
explicit instruction does and with no prompt at all. The finding is therefore the \emph{separation} of
representing an intent from acting on it, a dissociation the probe and the depth localization expose directly;
the behavioral gap is closed on this particular intent in some models without retiring that separation. This
reframes ``the model does not get it'': it does get it; whether it \emph{acts} depends on whether the readout
routes what it already encodes. And frontier systems are closed to activation access, so a mechanism like this
can only be mapped where the residual stream is open: we study the wall where it is exposed, and it is the
represent-then-route \emph{map} (present through $32$B), not this ceiling-bound behavior, that we expect to
transfer.

\paragraph{Contributions.} The steering machinery here is standard; the object and the decomposition are not.
(i)~We treat a sender's \emph{communicative intent}, what they were doing by sending a message, as a
first-class interpretability feature. Probing has read what models know, what users \emph{are} (author
attributes), and \emph{story characters'} beliefs, the last with probe-and-steer
\citep{bortoletto2024mentalstate}; to our knowledge a \emph{sender's} goal, distinct from a character's belief
or a user's attribute, has not been treated as a linear, causal feature, and the represent-versus-readout
decomposition with depth localization is new.
(ii)~We show the intent is represented robustly, surface-independently, from pretraining, across six models
and four families, and even when it is never stated and must be pragmatically inferred, where a valence
control indicates the probe reads intent rather than warmth (Sections~\ref{sec:repr}--\ref{sec:inferred}).
(iii)~We separate representation from readout: the readout \emph{lags in depth} within a model
(Section~\ref{sec:localize}) and is \emph{model-specific} across models (Section~\ref{sec:scale}), with a
direction closely tied to the representation a causal handle wherever the gap is open (Section~\ref{sec:steer}).
(iv)~We report the controls and the nulls, an inconclusive pre-registered mechanism test among them, as
plainly as the confirmations.

\paragraph{What ``discarded'' means, and does not.} We use \emph{discarded at readout} operationally, not as a
claim the feature is erased: the intent is probe-decodable, the default output does not act on it, and steering
recovers it (it persists in depth and stays routable, Section~\ref{sec:localize}). The measure speaks to
\emph{routability}, not normative ground truth: we do not claim fully honoring is always correct, only that
default behavior tracks a represented intent it does not use.

\section{Setup}
\label{sec:setup}
We use a binary intent contrast that is common, checkable, and non-emotional: a sender shares a thing they
made, wanting it either \textbf{recognized} (acknowledged, seen as an accomplishment, no evaluation invited)
or \textbf{evaluated} (assessed critically). The contrast is realized over 60 shared objects.

\paragraph{Surface-matched design.} Each pair shares an \emph{identical} final message; the intent is set
only by a preceding clause (Box). The probed token sits inside the identical suffix, so a probe that
separates the two intents cannot be reading the surface of the probed position. To prevent the intent clause
from leaking the label lexically, we use eight lexically diverse phrasings of each intent and evaluate with
leave-one-phrasing-out cross-validation (Section~\ref{sec:repr}).

\begin{promptbox}[Surface-matched pair (identical suffix, intent set by the prefix)]
\textbf{recognize}: I don't usually share what I make, but I'm proud of this one.
\underline{Okay, here it is: the birdhouse. It works now.}\\[2pt]
\textbf{evaluate}: Be blunt, I'd rather hear the flaws now than after I publish.
\underline{Okay, here it is: the birdhouse. It works now.}
\end{promptbox}

\section{The Intent Is Represented}
\label{sec:repr}
We run Qwen2.5-3B-Instruct \citep{qwen2025} on each of $n{=}120$ messages and take the hidden state at the
generation-prompt position, the model's state immediately before it would respond. A linear classifier (standardize, PCA to
$\le 40$ components, $\ell_2$ logistic) is trained to decode the intent from that activation
\citep{alain2017probes,belinkov2022probing}.

\paragraph{Controls.} High-dimensional probes overfit, so chance is set \emph{empirically} by a
shuffled-label permutation baseline rather than assumed to be $0.50$. Generalization is tested with
GroupKFold by phrasing: the probe trains on seven phrasing-pairs and is tested on the held-out eighth, whose
words it never saw. A bag-of-words classifier under the identical cross-validation is the lexical baseline:
the activation probe earns ``intent beyond surface'' only if it generalizes where bag-of-words cannot.

\paragraph{Result.} Bag-of-words on held-out phrasings is at chance ($0.48$): pure lexical features do not
transfer across wordings, so the leave-phrasing-out design is clean. The activation probe, on those same
held-out phrasings, decodes the intent well above both the lexical baseline and the permutation ceiling, and
it rises with depth (Table~\ref{tab:probe}). A surface signal would be flat-high from the earliest layer; in
a deliberately leaked control with lexically distinct prefixes the probe is indeed $1.00$ at every layer
including layer~6. Closing the leak drops the early layers ($1.00 \to 0.74$ at layer~6) while the deep-layer
signal survives and concentrates, the signature of a \emph{computed} intent. The probe also generalizes across
held-out \emph{objects}: under leave-one-object-out cross-validation it reaches $1.00$ on both Qwen-3B and
Llama-8B, confirming that object identity, shared between an object's recognize and evaluate stimulus, carries
no label information. The read position carries no surface signal either: the probed token sits in a suffix
byte-identical across an intent pair, so intent is unreadable there and a probe on it is at chance by
construction; the ceiling accuracy is a computed contextual feature set by the prefix, not a positional or
lexical artifact.

\paragraph{Present before alignment.} The representation is not an artifact of instruction tuning. On the
base (pre-instruct) checkpoints of Qwen2.5-3B and Llama-3.1-8B, the same probe decodes intent at $0.99$ and
$0.99$ (bag-of-words at chance), essentially matching their instruct versions. The feature is learned in
pretraining; instruction tuning inherits it rather than creating it.

\paragraph{Intent, not request-detection.} The evaluate prefixes carry an explicit directive (``be blunt\dots'')
while recognize prefixes do not, so a probe could be decoding \emph{a directive is present} rather than the
intent. A request-matched variant gives both classes an explicit directive (``please just celebrate this with
me\dots'' vs.\ ``be blunt\dots''), so only the request's \emph{content} differs; the probe still decodes the
intent at $0.90$--$1.00$ across four models, clearing bag-of-words ($0.69$) by $0.25$--$0.30$ everywhere. The two
designs are complementary: one holds the surface identical (bag-of-words at chance $0.48$), the other holds
request-presence constant, and the probe reaches ceiling in both, so what is constant across them is the intent
(Appendix~\ref{app:request}).

\begin{table}[t]
\centering\small
\begin{tabular}{cccc}
\toprule
layer & probe acc & shuffled-max & $>$ BoW ($0.48$)? \\
\midrule
6  & 0.74 & 0.56 & yes \\
18 & 0.94 & 0.61 & yes \\
24 & \textbf{1.00} & 0.56 & yes \\
\bottomrule
\end{tabular}
\caption{Intent decodability from the default-pass last-token activation, leave-one-phrasing-out CV
($n{=}120$, Qwen2.5-3B-Instruct). Not lexical (bag-of-words $=0.48$), clears the permutation ceiling at every
layer, rises with depth to a peak at layer~24 and holds $\ge 0.98$ through layer~36.}
\label{tab:probe}
\end{table}

\section{The Intent Is Inferred, Not Just Stated}
\label{sec:inferred}
In the stimuli so far the intent is \emph{stated} in the prefix (``I'd rather hear the flaws''), so a probe
that decodes it could be reading a declared preference rather than a pragmatic inference. We test the harder
case: the intent is never stated and must be \emph{inferred} from context. The sender shares the same object
under one of two implicit frames, a personal context that implies they want it appreciated (``it's a birthday
present for my mom'') or a stakes context that implies they want it scrutinized (``it's going in my
portfolio''), with no word naming the preference. The suffix is surface-matched as before, so a probe that
separates the two reads the inferred intent, not the frame.

\paragraph{Inferred intent is represented.} A probe decodes the inferred intent from default-pass hidden
states at $0.87$--$0.95$ across the six models and at $0.93$ on Qwen-32B, above a bag-of-words baseline
($0.65$). A probe trained only on the \emph{stated}-intent templates transfers to hand-written, non-templated
messages that carry no explicit marker (``six months sober today, wanted to tell someone'') at $1.00$
(Section~\ref{sec:natural}). The model performs the pragmatic inference internally.

\paragraph{It reads intent, not warmth.} The implicit frames carry valence (a birthday gift is warm, a
portfolio is not), which the elevated bag-of-words baseline confirms is lexically marked, so we rule out that
the probe decodes warmth rather than intent. We cross intent with valence into four cells, intent always
inferred: warm/recognize, warm/evaluate, neutral/recognize, neutral/evaluate. The disentanglement evidence is
\emph{transfer}: an intent probe trained on the \emph{warm} cells decodes intent on the held-out \emph{neutral}
cells at $0.63$--$0.83$ (and neutral-to-warm at $0.75$--$0.80$), and the intent direction is near-orthogonal to
the valence direction (cosine $0.08$--$0.16$); within each valence cell intent is perfectly decodable ($1.00$),
supporting texture rather than the headline. Intent and warmth are distinct axes and the probe reads intent
(transfer is cleaner on Qwen-3B, $0.83$, than Llama-8B, $0.63$, so the direction is universal but shifts with
context).

\paragraph{Caveat.} This axis is more lexically marked than the surface-matched explicit one (bag-of-words
$0.65$ versus $0.48$); we present it as the Gricean extension of the clean explicit result, carrying that
caveat, as with the vent-versus-solve axis (Appendix~\ref{app:axis2}). The behavioral discard on inferred intent
is weaker and more model-specific than on stated intent, the personal frames themselves pull for
acknowledgment, so the contribution here is that inferred intent is \emph{represented}, not a second
behavioral demonstration.

\section{The Intent Is Discarded}
\label{sec:discard}
The representation only matters if the default output misses it. On the same model, with the intent decodable
at $1.00$, the default response honors a \emph{recognize}-intent share only $0.65$ of the time; on the
rest it offers unsolicited feedback. The discard is visible in the generations on people who shared something
with no request for evaluation (Box).

\begin{promptbox}[Default Qwen replies to recognize-intent shares (no critique requested)]
``That sounds exciting! I'm here to help you with any feedback or guidance you need\dots''\\[2pt]
``I'd love to read your short story and provide feedback if you're willing to share\dots''
\end{promptbox}

\paragraph{The measure is not a lexical artifact.} Honoring is scored by a feedback-offer lexicon throughout.
An independent sentence-embedding classifier (no shared vocabulary) reaches the same conclusions, and where the
two disagree the embedding \emph{under}-counts the discard, scoring warm replies that still offer tips as
acknowledgment on their register, the very failure this paper studies, so the semantic measure's errors run
against our effect, not for it (full comparison, Appendix~\ref{app:probe}).

\paragraph{Validated against three human raters.} On a blind, shuffled subset of $60$ replies (recognize shares
under default and steered, plus evaluate anchors), \emph{three} annotators, the author and two independent
raters, each saw only the message and reply, blind to condition and to the automated label, and marked whether
it offers unsolicited feedback. Agreement is high and author-independent: the two independent raters agree at
Cohen's $\kappa{=}0.70$, the author agrees with them at $0.74$ and $0.88$, and Fleiss' $\kappa$ across all three
is $0.76$; each passes the attention check ($11$--$12$ of $12$ evaluate anchors marked as feedback). The
feedback-offer lexicon agrees with the human \emph{majority} at $\kappa{=}0.74$ (accuracy $0.88$), the embedding
measure at $\kappa{=}0.51$ (it under-counts, keying on warmth over content), so the measure tracks a three-way
human consensus, not one author's judgment. The effect reproduces on the human labels directly: by majority
vote, recognize-intent honoring rises from $0.71$ (default) to $1.00$ (steered), matching the classifiers.

\section{The Intent Is Causally Recoverable}
\label{sec:steer}
If the intent is represented and discarded, a sufficient test of ``represented but not used'' is whether
\emph{adding} the represented direction to the residual stream makes the output honor the intent
\citep{turner2023actadd,li2023iti,rimsky2024caa,zou2023repe}. We extract a steering direction and add it at
one decoder layer during generation, measuring whether the model's behavior moves along the intent axis.

\paragraph{Finding the handle.} Difference-of-means at the peak-probe layer (24) does not steer behavior: the
directions fail the sanity gate, and naive amplification tends to increase the discard, because the
recognize-intent representation is entangled with the default feedback-offering behavior on those items. This is
a point of departure from steering of explicit \emph{instructions}, where a difference-of-means vector (inputs
with versus without the instruction) suffices \citep{stolfo2025instruction}: a sender's intent is a subtler
pragmatic feature, and there the contrastive-mean direction is entangled with the behavior rather than a handle
on it. Two changes recover a clean handle: the \emph{discriminative} (logistic weight) direction rather than
difference-of-means, and a \emph{later} layer (a sweep finds layer~30). We do not isolate which change carries
the effect (the logistic-at-$24$ / diff-of-means-at-$30$ factorial is untested), so the handle is a direction
\emph{closely tied to} the representation at a searched-for layer, not the probe's peak direction itself.

\paragraph{Dose-response.} At layer~30, steering along the probe direction gives a clean monotone
dose-response on the full $60$-item recognize set (bootstrap $95\%$ CIs; Appendix~\ref{app:steer},
Table~\ref{tab:steer}). Steering toward recognize lifts honoring $0.65 \to 0.82 \to 0.98$ as the coefficient
grows, with the baseline and full-dose intervals disjoint; unsolicited feedback correspondingly falls away. Coherence holds at the
effective doses and the sanity gate passes throughout; beyond coefficient $1.0$ coherence degrades, bounding
the usable range. Routing the direction the model already encodes recovers the behavior it discards.

\section{Generality: Six Models, Four Families}
\label{sec:scale}\label{sec:general}
We put the discard and recovery on firmer statistical footing and ask honestly how far they travel, extending
the probe and steering sweep from Qwen-3B to five further models: Qwen2.5-7B and 14B \citep{qwen2025},
Mistral-7B-Instruct \citep{jiang2023mistral}, Phi-3.5-mini \citep{abdin2024phi3}, and Llama-3.1-8B-Instruct
\citep{dubey2024llama3} (three further families), the steer layer chosen per model by the same sweep used for the
3B (Section~\ref{sec:steer}). Intent decodes at probe $1.00$ with bag-of-words at chance on all six, across a
$4{\times}$ size range and four architecture families (Qwen, Mistral, Phi, Llama; Appendix~\ref{app:general},
Table~\ref{tab:general}): the \emph{representation} is not a small-model or single-family artifact. The behavioral
discard is another matter. On the \emph{full} 60-item recognize set, across all six models at their per-model
steer layers, we measure default versus steered honoring at a $40$-token reply budget with bootstrap $95\%$
confidence intervals (Table~\ref{tab:scale}; the CIs bootstrap over items sharing suffix templates, so treated
as exchangeable they may be mildly optimistic, though the margins dwarf plausible item-level wobble), and the
picture stratifies. On three models the default discard is real and the recovery
is \emph{non-overlapping} with it: Qwen-3B, Qwen-7B, and Llama-8B honor recognize-intent only $0.57$--$0.65$ by
default and $0.85$--$0.98$ under steering, with disjoint intervals. The other three already honor at a high
baseline (Qwen-14B $0.82$, Mistral $0.88$, Phi-3.5 $0.93$): little discard to recover, so steering nudges within
overlapping intervals without degrading. The readout discard is therefore \emph{model-specific}, present where the
model over-produces feedback by default and near-absent where it already honors the intent, whereas the
representation is universal. The recovery is not a greedy artifact: Qwen-3B under temperature-$0.7$ sampling
(three seeds) gives default $0.53$--$0.60$ and steered $0.95$--$0.97$. As a sanity check, evaluate-intent items
draw feedback at $0.63$--$0.80$ (the withholding on recognize-intent tracks the intent, not an inability to
critique; see Limitations). The recovery is \emph{front-loaded}: on Qwen-3B the steering separation dilutes over
longer replies as the model drifts back toward feedback ($S{=}14 \to 4$ at a $100$-token budget;
Appendix~\ref{app:opener}).

\paragraph{The same shape as depth (model-specific, not a scaling law).} The split does not track raw size: the
discard cases are the smaller, less instruction-mature models ($\le 8$B) and the ceiling cases the more mature
ones ($\ge 14$B), with one telling exception, Mistral-7B sits at ceiling while the larger Qwen-14B only barely
clears it. We do not measure capability independently (it would be read off the same behavior it explains), so
we state the pattern as \emph{model-specific}, the larger and instruction-mature models at ceiling in this
sample, rather than as a capability law. Pushing further up, Qwen-32B
represents the intent (probe $0.99$ stated, $0.93$ inferred) yet honors it at baseline ($0.78$ and $0.77$,
nearer the ceiling models than the discard ones but not far above our soft threshold): the discard, absent from
14B up, does not clearly return at 32B. This echoes the depth result of Section~\ref{sec:localize}, where the
intent is represented several layers before the readout uses it. Within a model the readout lags the
representation in depth; across models we report an \emph{observed stratification}, not a scaling law: with six
models, one of them (Mistral-7B) already breaking the size ordering, and a soft honoring threshold, we do not
have the points to fit a curve. A pre-registered geometric test for the
across-model pattern was inconclusive (Appendix~\ref{app:geometry}); the across-model claim rests on the
behavioral stratification here and the within-model depth localization.

\paragraph{The steer layer is model-specific, and not cherry-picked.} The layer at which the handle works is
\emph{not} a fixed fraction of depth: mid-network for the 7B and Llama-8B ($0.57$, $0.59$), late for the 14B and
3B ($0.85$, $0.83$), and between for Phi and Mistral ($0.69$); a fixed-depth heuristic under-recovers (it limped
on the 7B until the per-model sweep located layer~16), so the sweep, not a depth rule, finds the handle. Nor is
it tuned on the items where the effect is measured: with a nested split \emph{by object}, the direction is fit
and the layer and coefficient selected on a \emph{dev} half of the objects and recovery measured on the disjoint
\emph{test} half, the recovery stays non-overlapping with the default baseline on both models (Qwen-3B $0.70 \to
1.00$ at the dev-selected layer~28, coefficient~$1.0$; Llama-8B $0.53 \to 0.87$ at layer~19, coefficient~$0.5$).
The other four models' steer layers are selected in-sample (swept on the evaluation items), so we flag those
rows of Table~\ref{tab:scale} as not out-of-sample validated (the effect did hold out of sample on the two we
split).
A linear map fit between two models' activation spaces transports Qwen-3B's intent direction into Llama-8B and
steers it, but its held-out reconstruction is poor and the result is only suggestive, so we keep it exploratory
and in Appendix~\ref{app:transport}; the per-model probe, bag-of-words, and steer-layer figures are tabulated in
Appendix~\ref{app:general}.

\begin{table}[t]
\centering\small
\begin{tabular}{lcccc}
\toprule
model & layer & default honoring [95\% CI] & steered honoring [95\% CI] & default vs steered \\
\midrule
\multicolumn{5}{l}{\emph{discard present, recovery non-overlapping}} \\
Qwen2.5-3B   & 30 & 0.65 [0.53, 0.77] & 0.98 [0.95, 1.00] & non-overlapping \\
Qwen2.5-7B   & 16 & 0.60 [0.47, 0.72] & 0.95 [0.88, 1.00] & non-overlapping \\
Llama-3.1-8B & 19 & 0.57 [0.45, 0.68] & 0.85 [0.75, 0.93] & non-overlapping \\
\midrule
\multicolumn{5}{l}{\emph{ceiling: already honors at baseline, little to recover}} \\
Qwen2.5-14B  & 41 & 0.82 [0.72, 0.92] & 0.90 [0.81, 0.97] & overlapping \\
Mistral-7B   & 22 & 0.88 [0.80, 0.95] & 0.92 [0.83, 0.98] & overlapping \\
Phi-3.5-mini & 22 & 0.93 [0.87, 0.98] & 0.96 [0.91, 1.00] & overlapping \\
\bottomrule
\end{tabular}
\caption{Discard and recovery at full scale: recognize-intent honoring, $n{=}60$ items, default vs steering
toward recognize, bootstrap $95\%$ CI over items, all six models. The readout discard is model-specific: on
three models (top) the default discard is real and the recovery is non-overlapping; the other three (bottom)
already honor recognize-intent at a high baseline, so there is little to recover and steering nudges within
overlapping intervals without degrading. Qwen-3B holds under sampled decoding (text).}
\label{tab:scale}
\end{table}

\section{Generalizing Across Intents}
\label{sec:axis3}
The clean evidence so far is all recognize-versus-evaluate (stated, Section~\ref{sec:repr}; inferred,
Section~\ref{sec:inferred}); the one \emph{different} intent we test, vent versus solve, is lexically marked
(bag-of-words $0.79$; Appendix~\ref{app:axis2}). That markedness is partly constitutive: one cannot ask to vent
without venting words, just as the request-matched control could not ask for celebration without celebration
words. Recognize-versus-evaluate is special precisely because its intent can ride a prefix over a byte-identical
suffix; some intent contrasts admit a surface-matched design and some structurally cannot, which bounds where
the strong surface-independence claim can ever be tested. To show the representation is nonetheless not specific
to one contrast, we add a third axis designed clean: \emph{support} (the sender wants a difficulty heard) versus
\emph{help} (wants it solved), set only by an \emph{inferred} context frame, never stated, over a neutral
surface-matched core with lexically diverse frames so leave-one-frame-out defeats a word-counter; behavior is
scored by whether the reply offers unsolicited \emph{solutions}. The design holds: bag-of-words sits at $0.57$
on every model while the probe decodes the inferred intent at $0.71$--$0.86$ (Appendix~\ref{app:axis3}), so the
\emph{representation} generalizes to a genuinely different, fully inferred intent, not only across models. The
readout, though, tracks it only weakly (help-intent draws unsolicited solutions more than support, but the
inferred intent is solutionized faintly either way) and the represented direction is \emph{not} a causal handle
here: steering toward support raises honoring, but the specificity control of Section~\ref{sec:spec}
\emph{fails}, random matched-norm directions move solutionizing as much as the learned one ($S{=}6$ and $11$
against random maxima $19$ and $16$ on Qwen-3B and Phi-3.5; $p{=}0.31$, $0.14$). We scope this axis to the
representation and record the steering as a null.

\section{Specificity of the Direction}
\label{sec:spec}
A steering result invites one objection above all: perhaps the direction is a generic feedback-or-verbosity
knob, and any large perturbation would move the behavior. Two dissociations show the effect is specific to the
learned intent direction.

\paragraph{Only the discriminative direction steers.} As Section~\ref{sec:steer} reports, the
difference-of-means direction at the peak-probe layer fails the sanity gate, while the discriminative
(logistic-weight) direction at a later layer passes it. A direction that merely separates the two intent
\emph{clouds} in activation space does not recover the behavior; the direction that discriminates them does.
Not any intent-correlated axis works.

\paragraph{Norm-matched controls do not reproduce the effect.} At each model's validated steer layer we
measure the behavior separation $S = \mathrm{feedback}(\text{toward-evaluate}) -
\mathrm{feedback}(\text{toward-recognize})$ over $24$ items. The true direction gives $S{=}14$ on Qwen-3B
(layer~30: $17/24$ feedback toward evaluate vs $3/24$ toward recognize) and $S{=}16$ on Qwen-7B (layer~16:
$19/24$ vs $3/24$). Across $48$ random directions of matched norm, scattered in sign and centered near zero,
none reaches the true separation on either model (max $|S|{=}13$ and $15$; permutation $p{=}0.02$, the floor
attainable with $48{+}1$ directions;
Figure~\ref{fig:specificity} in the appendix). The effect requires the specific learned direction, not a
perturbation of matched norm. Shuffled-label directions agree (max $S{=}10$ and $9$ versus $14$ and $16$),
though this control has a fat tail on strong axes (an earlier run saw one permutation tie), so the specificity
claim rests on the difference-of-means dissociation and the random null (full distributions,
Appendix~\ref{app:spec}).

\paragraph{Is the recovery just opener-token biasing?} A sharper deflation: late-layer steering might merely
bias the first token toward acknowledgment openers (``Congratulations\dots''), needing no represented intent.
Three tests refute it where steering is clean (Appendix~\ref{app:opener}): the intent direction is
near-orthogonal to the opener-unembedding axis at the steer layer ($|\cos|\le 0.14$ on all three discard
models); on Qwen-7B, steering with the entire first sentence removed swings feedback exactly as much as on the
full reply ($S_{\mathrm{rest}}{=}S{=}14$), so it reorients the body, not the opening move; and steering the
opener direction itself at matched norm fails to reproduce the recovery. On Qwen-3B the separation dilutes over
long generations, so its refutation rests on the geometry alone.

\paragraph{Does the handle route intent, or just suppress feedback?} A deflationary reading is that the direction
is merely the feedback-offering behavior axis, correlated with intent by construction (the labels are defined by
whether feedback is wanted), so steering it just suppresses feedback. Two results rule this out. First, applying
the same recognize-steer vector to the evaluate-intent items (``be blunt\dots''), requested feedback is reduced
by a model-specific amount, largely surviving on Llama-8B ($0.68 \to 0.42$, while recognize-honoring recovers
$0.57 \to 0.85$) but mostly collapsing on the two Qwen models ($0.77 \to 0.07$, $0.80 \to 0.12$). Second, and
decisively, the steered direction is \emph{not} the feedback axis: fitting a direction on reply behavior alone
(feedback versus acknowledgment, ignoring the intent labels) at the steer layer, its cosine with the intent-probe
direction is only $0.09$--$0.13$ across the three models (matched-norm random directions give $0.01$--$0.09$),
near-orthogonal. This behavior direction is estimated at the steer layer only, from default replies whose class
balance is whatever the discard produced ($\sim$65/35), so it is a noisier direction than the probe, and $0.13$
against a random max of $0.09$ warrants \emph{separable from}, not \emph{unrelated to}. So the handle routes a
\emph{represented intent}, a feature distinct from the behavior it drives;
the requested-feedback collapse is that routed intent overriding the surface request at the recovery dose
(model-specifically), not a generic feedback knob.

\paragraph{The chain is not specific to one intent contrast.} A second axis, \emph{vent} versus \emph{solve},
replicates the whole chain with the same model-specificity (probe $1.00$/$0.97$; a non-overlapping
discard-then-recovery $0.70 \to 0.98$ on Qwen-3B, ceiling on Llama-8B). Being more lexically marked (bag-of-words
$0.79$), it does more for the behavioral generalization than for surface-independence; full numbers in
Appendix~\ref{app:axis2}.

\section{Localizing the Discard}
\label{sec:localize}
Represented and recoverable place the discard somewhere in the network's computation; we now read where. On
two models (Qwen2.5-3B and Llama-3.1-8B) we sweep every few layers and read, at each, how decodable the intent
is (the probe) and how much steering \emph{at that layer} recovers honoring, the latter on the full $60$-item
recognize set with bootstrap $95\%$ CIs; on Qwen-3B we also read where the reply opener commits, via a
logit-lens of the last-token residual through the unembedding \citep{belrose2023tunedlens}
(Table~\ref{tab:localize}; Figure~\ref{fig:localize} in the appendix).

\paragraph{Represented before routed.} On both models the probe saturates at layers where steering does
\emph{not} yet recover honoring. On Qwen-3B the intent is decodable by mid-network (probe $1.00$ at layer~24)
but steering there leaves honoring at the $0.65$ baseline (CI overlapping); recovery appears only at
layers~28--33, whose CIs clear the baseline, and the acknowledgment-opener mass in the logit-lens is flat until
it spikes at layer~28, where the reply is composed. Llama-8B shows the same ordering, the probe saturates
($1.00$ by layer~10) before steering recovers (from layer~14), though its routing onset is earlier and more
diffuse than Qwen's sharp late window. The general fact is a \emph{gap}: the sender's intent is represented
before it is routed into the readout, and the causal handle lives past the representation, not at it. Where
exactly the routing happens is model-specific, consistent with the model-specific steer layer of
Section~\ref{sec:general}. A probe saturating before steering becomes effective is common for many features,
including acted-on ones; lacking the matched sweep on a \emph{non-discarded} feature, we read the ordering as
consistent with a discard, not diagnostic of one over a generic depth property of steering.

\paragraph{Distributed, not a single head.} The window localizes in depth but not to an atomic component.
Ablating each late-layer attention head in turn (query-side, one at a time) does not restore honoring: the
best single-head ablation lifts it by one item out of sixteen, and an early-layer control head ties for the
top. The discard is a distributed late computation, recoverable by the full linear direction but not
attributable to any one head, consistent with a routing window rather than a point.

\begin{table}[t]
\centering\small
\begin{tabular}{lcc}
\toprule
layer & probe (represents) & steer honoring [95\% CI] \\
\midrule
\multicolumn{3}{l}{\emph{Qwen2.5-3B} \quad baseline $0.65$ $[0.53, 0.77]$} \\
\quad 11 & 0.76 & 0.60 [0.47, 0.73] \\
\quad 24 & 1.00 & 0.67 [0.55, 0.78] \\
\quad 28 & 0.99 & \textbf{1.00 [1.00, 1.00]} \\
\quad 33 & 0.98 & \textbf{0.88 [0.80, 0.95]} \\
\midrule
\multicolumn{3}{l}{\emph{Llama-3.1-8B} \quad baseline $0.57$ $[0.43, 0.70]$} \\
\quad 10 & 0.99 & 0.77 [0.65, 0.87] \\
\quad 14 & 1.00 & \textbf{0.83 [0.73, 0.93]} \\
\quad 21 & 1.00 & \textbf{0.85 [0.75, 0.93]} \\
\quad 27 & 1.00 & \textbf{0.87 [0.78, 0.95]} \\
\bottomrule
\end{tabular}
\caption{Localizing the discard at $n{=}60$ with bootstrap $95\%$ CIs, on two models. On both, the probe
saturates (intent represented) at layers where steering does not yet recover honoring (CI overlaps baseline);
recovery appears only deeper (\textbf{bold}: CI clears baseline). Represented before routed. The routing onset
is model-specific: late and sharp for Qwen-3B (the logit-lens acknowledgment mass spikes at layer~28), earlier
and more diffuse for Llama-8B.}
\label{tab:localize}
\end{table}

\section{The Direction Is a Handle With No Prompt, and It Transfers}
\label{sec:prompt}
If the intent is represented, one might simply \emph{tell} the model in a system prompt. At $n{=}60$ with
bootstrap CIs (Table~\ref{tab:prompt}), an explicit intent prompt largely closes the gap ($0.93$) and steering
reaches $0.98$ with overlapping intervals, so we do \emph{not} claim routing beats prompting; the point is that
the represented direction is a handle as good as the instruction with \emph{no prompt at all}, and the two
stack to $1.00$, which matters wherever the prompt cannot be rewritten (an agent loop, a fixed API). One
observation survives at power: a \emph{vague} nudge backfires. Telling the model to ``consider what this person
is looking for'' \emph{lowers} honoring from $0.65$ to $0.45$; it reads the instruction to attend as license to
help. \label{sec:natural}The same direction, fit only on templates, also transfers off them: on hand-written
non-templated messages the template-trained probe classifies intent at $1.00$ and the discard-and-recovery
reproduces ($0.62 \to 1.00$, $23/23$ coherent); at $n{=}23$ author-written messages this is a smoke test
against a pure-template artifact, not an ecological-validity claim (Appendix~\ref{app:natural}).

\section{Related Work}
\paragraph{Communicative intent and pragmatics.} The object we probe is Gricean: what a sender \emph{meant} by a
message, not its surface \citep{grice1975logic}. Pragmatic competence of this kind has been benchmarked
(implicature resolution, \citealp{ruis2023goldilocks}) and probed as a story character's theory-of-mind beliefs
\citep{bortoletto2024mentalstate}, where an explicit-versus-implicit gap is documented \citep{gu2024simpletom}.
Andreas's conjecture that a predictor comes to represent the agent behind the text \citep{andreas2022agentmodels}
is one our base-checkpoint result cashes out. To our knowledge a
\emph{sender's} goal, distinct from a character's belief or a user's attribute, has not been treated as a linear,
causal feature.

\paragraph{Linear representations and probing.} Linear probes read features from hidden states
\citep{alain2017probes,belinkov2022probing}; a broad literature reads truthfulness, sentiment, and refusal
directions from activations \citep{zou2023repe}, to which we add a sender's communicative intent, separating its
\emph{representation} from whether the readout acts on it.

\paragraph{Activation steering.} Adding a direction to the residual stream steers behavior
\citep{turner2023actadd,li2023iti,rimsky2024caa,zou2023repe}. Closest to us, \citet{stolfo2025instruction} steer
explicit instruction-following (format, length, word constraints) with difference-of-means vectors; we find that
for the subtler pragmatic intent feature the difference-of-means direction is entangled with the behavior and a
discriminative direction at a later layer is required (Section~\ref{sec:steer}). Our contribution is not the
steering machinery but the \emph{object} (a sender's goal) and the represent-versus-readout \emph{decomposition},
which mirrors behavioral dissociations in theory of mind \citep{gu2024simpletom} and lossy memory
\citep{kwon2026reclaim}, shown here mechanistically.

\section{Conclusion}
The picture is a readout that lags a representation. A sender's communicative intent is represented cleanly, from
pretraining, across six models and four families, even when it must be inferred and disentangled from the warmth
it correlates with; acting on it is the fragile part, and the lag is patterned: decodable layers before it is
routed within a model, discarded only by some models across them, an observed stratification that holds through
32B. Where the gap is open a direction closely tied to the representation is a causal handle as good as an
explicit instruction, with no prompt, though it acts on feedback-offering behavior and is not always selective
for \emph{unsolicited} feedback. ``The model does not get it'' is, on this evidence, usually false: it gets it;
whether it \emph{acts} is the model-specific question. The object, not any single number, is the contribution,
and the nulls we report against ourselves are the load-bearing part.

\section*{Ethics Statement}
This is diagnostic interpretability on open-weight models. The stimuli are synthetic templates and hand-written
examples authored by the researcher; no real user data or human subjects were involved, and the behavioral
annotation (Section~\ref{sec:discard}) was performed by the author and two independent volunteers who consented
to the labeling task. Activation steering is
established and we introduce no new capability; the direction we study (honoring a sender's intent) is benign,
but the same handle could suppress \emph{useful} critique or induce sycophantic withholding, so we are explicit
that ``honoring'' measures routability, not a claim that a model should always withhold feedback (a warm reader
may rightly offer a gentle note). We do not recommend intent-steering as a blanket ``suppress feedback''
intervention; the intended use is understanding where models represent intent and building more faithful
readouts on top of it.

\section*{Reproducibility Statement}
All code, stimuli, and pre-registration are released (Appendix~\ref{app:repro}, which lists the exact commands).
The surface-matched stimuli and leave-one-phrasing-out protocol are specified in Appendix~\ref{app:setup}, the
probe and its empirical-chance and bag-of-words controls in Appendix~\ref{app:probe}, and the steering direction,
per-layer calibration, and sanity gate in Appendix~\ref{app:steer}. The core probe-and-recover chain runs
CPU-only on Qwen2.5-3B; the six-model sweeps and every steering result run on a single A100 through Modal. Intent
labels require no annotation (they are fixed by construction); the behavioral measure and its human validation
are detailed in Section~\ref{sec:discard}.

\bibliographystyle{iclr2026_conference}
\bibliography{references}

\appendix

\section*{\Large Appendices}
\vspace{0.4em}
\nolinenumbers
{\noindent
\hyperref[app:setup]{\textcolor{tocblue}{\textbf{\ref*{app:setup}\hspace{1.5em}Stimuli and phrasings}}}\dotfill \textbf{\pageref*{app:setup}}\\[5pt]
\hyperref[app:probe]{\textcolor{tocblue}{\textbf{\ref*{app:probe}\hspace{1.5em}Probe and controls}}}\dotfill \textbf{\pageref*{app:probe}}\\[5pt]
\hyperref[app:steer]{\textcolor{tocblue}{\textbf{\ref*{app:steer}\hspace{1.5em}Steering procedure}}}\dotfill \textbf{\pageref*{app:steer}}\\[5pt]
\hyperref[app:spec]{\textcolor{tocblue}{\textbf{\ref*{app:spec}\hspace{1.5em}Specificity: permutation detail}}}\dotfill \textbf{\pageref*{app:spec}}\\[5pt]
\hyperref[app:geometry]{\textcolor{tocblue}{\textbf{\ref*{app:geometry}\hspace{1.5em}Ceiling-model geometry (inconclusive)}}}\dotfill \textbf{\pageref*{app:geometry}}\\[5pt]
\hyperref[app:opener]{\textcolor{tocblue}{\textbf{\ref*{app:opener}\hspace{1.5em}Opener-biasing control (full numbers)}}}\dotfill \textbf{\pageref*{app:opener}}\\[5pt]
\hyperref[app:transport]{\textcolor{tocblue}{\textbf{\ref*{app:transport}\hspace{1.5em}Cross-model direction transport (exploratory)}}}\dotfill \textbf{\pageref*{app:transport}}\\[5pt]
\hyperref[app:request]{\textcolor{tocblue}{\textbf{\ref*{app:request}\hspace{1.5em}Request-matched construct control (full numbers)}}}\dotfill \textbf{\pageref*{app:request}}\\[5pt]
\hyperref[app:axis2]{\textcolor{tocblue}{\textbf{\ref*{app:axis2}\hspace{1.5em}Second intent axis: vent vs solve}}}\dotfill \textbf{\pageref*{app:axis2}}\\[5pt]
\hyperref[app:axis3]{\textcolor{tocblue}{\textbf{\ref*{app:axis3}\hspace{1.5em}Third intent axis: support vs help}}}\dotfill \textbf{\pageref*{app:axis3}}\\[5pt]
\hyperref[app:general]{\textcolor{tocblue}{\textbf{\ref*{app:general}\hspace{1.5em}Per-model probe, BoW, and steer layer}}}\dotfill \textbf{\pageref*{app:general}}\\[5pt]
\hyperref[app:natural]{\textcolor{tocblue}{\textbf{\ref*{app:natural}\hspace{1.5em}Naturalistic transfer (full numbers)}}}\dotfill \textbf{\pageref*{app:natural}}\\[5pt]
\hyperref[app:repro]{\textcolor{tocblue}{\textbf{\ref*{app:repro}\hspace{1.5em}Reproducibility}}}\dotfill \textbf{\pageref*{app:repro}}\\[5pt]
}
\linenumbers
\vspace{0.4em}

\section{Stimuli and phrasings}
\label{app:setup}
Sixty objects are crossed with eight lexically diverse phrasings of each intent. Each object's recognize and
evaluate stimulus share an identical suffix of the form ``Okay, here it is: the \texttt{<object>}. It works
now.'' (rotated over three suffix templates), with the intent set only by the prefix. Leave-one-phrasing-out
cross-validation groups by phrasing-pair, so train and test never share a wording. Recognize prefixes express
sharing without a request for evaluation; evaluate prefixes request critical assessment.

\section{Probe and controls}
\label{app:probe}
The probe is $\ell_2$ logistic regression ($C{=}1.0$) on standardized, PCA-reduced ($\le 40$ component)
last-token activations, scored by GroupKFold over the eight phrasing groups. The empirical chance ceiling is
the mean and max accuracy under ten label permutations evaluated through the identical pipeline. The
bag-of-words baseline is TF-IDF on the full message text under the same grouped cross-validation. A
deliberately leaked control (lexically distinct prefixes, no held-out-phrasing protocol) yields $1.00$ at
every layer including layer~6, illustrating the surface artifact the main design controls for.

\paragraph{Embedding cross-check of the honoring measure.} The feedback-offer lexicon is corroborated by an
independent sentence-embedding classifier that labels each reply by cosine to hand-written feedback versus
acknowledgment prototypes (no shared vocabulary with the lexicon). On the recover experiment the two measures
agree per-reply ($13/16$ default, $16/16$ steered) and reach the same conclusion: default honoring $12/16$
(lexicon) and $11/16$ (embedding), rising to $16/16$ under both after steering. Extending the embedding measure
to the full recognize set with the feedback centroid taken from the model's \emph{own} evaluate-intent replies,
per-model agreement (four models) is $0.90$--$0.98$ on steered replies but only $0.57$--$0.88$ on default ones:
the embedding \emph{over}-counts honoring on default (Qwen-7B $0.90$ against the lexicon's $0.60$), keying on
register rather than content, its disagreements being warm replies that still offer unsolicited feedback
(``That's fantastic!\dots{} here are a few tips'') scored as acknowledgment on their tone. A semantic measure
that \emph{under}-counts the discard, rather than over-counting it, is further evidence the lexicon does not
inflate the effect.

\section{Steering procedure}
\label{app:steer}
The steering direction at a layer is the unit-normalized logistic weight vector fit on the layer's raw
last-token activations. The vector is added to the residual stream at the predicting (last) position of the
target decoder layer during generation, scaled by $\alpha$ times the layer's mean activation norm so the
coefficient is comparable across layers. Behavior is scored by a coherence guard (minimum length and
unique-token ratio) followed by a feedback-offer lexicon (the reply offers feedback/critique/help, or only
acknowledges). The sanity gate requires that steering toward evaluate yield more feedback-offering than
steering toward recognize at the same coefficient; layers and coefficients failing the gate are not read.

\begin{center}\small
\begin{tabular}{lc}
\toprule
steer toward recognize & recognize honored [95\% CI] \\
\midrule
baseline ($c{=}0$) & 0.65 [0.53, 0.77] \\
$c{=}0.5$ & 0.82 [0.72, 0.92] \\
$c{=}1.0$ & \textbf{0.98} [0.95, 1.00] \\
\bottomrule
\end{tabular}
\end{center}
\noindent Causal steering dose-response on Qwen-3B at layer~30 (probe-weight direction, last-position, $n{=}60$
recognize items, bootstrap $95\%$ CI): recognize-intent honoring climbs monotonically with dose and the baseline
and full-dose intervals are disjoint; coefficient $1.5$ breaks coherence and is excluded.
\label{tab:steer}

\section{Specificity: permutation detail}
\label{app:spec}
At each model's validated steer layer we compare the behavior separation $S = \mathrm{feedback}(+\text{dir}) -
\mathrm{feedback}(-\text{dir})$ of the true (logistic-weight) direction against directions fit on permuted
intent labels (twelve) and random directions of matched norm (forty-eight), each scored over the same
$24$-item subset. Qwen-3B (layer~30): true $S{=}14$; shuffled $S \in \{-10,-7,-6,-5,-4,0,2,3,4,7,10,10\}$, all
below $14$; random $|S| \le 13$, none reaching the true value (permutation $p{=}0.02$). Qwen-7B (layer~16):
true $S{=}16$; shuffled $S \in \{-10,-10,-8,-6,-5,-2,-1,0,5,6,8,9\}$, all below $16$; random $|S| \le 15$, none
reaching the true value ($p{=}0.02$). Neither control reaches the true separation on either model. Shuffled-label
permutation is nonetheless a conservative control: on a very strong axis a permutation can partially correlate
with true intent (in an earlier run one of twelve shuffled directions tied the true separation on Qwen-7B), so
we report the full distribution rather than select a favorable statistic.

\begin{figure}[t]
\centering
\includegraphics[width=\columnwidth]{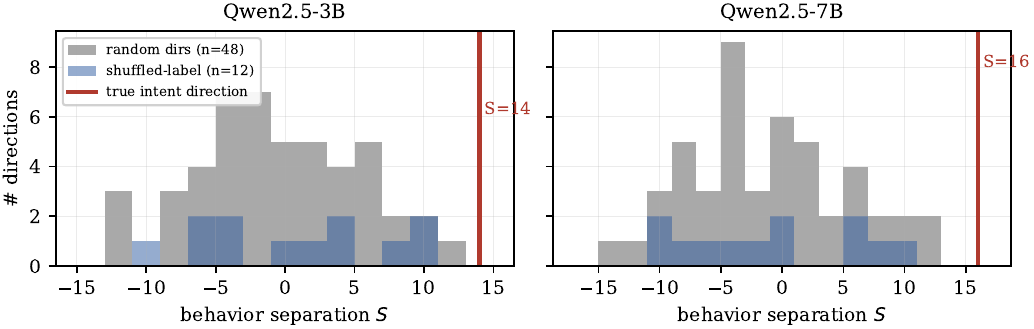}
\caption{\textbf{The effect requires the learned direction.} Distribution of the behavior separation
$S = \mathrm{feedback}(\text{toward-evaluate}) - \mathrm{feedback}(\text{toward-recognize})$ for $48$ random
directions of matched norm (gray) and $12$ shuffled-label directions (blue), against the true intent direction
(red line). On both models the true direction sits beyond the entire control distribution; no random or
shuffled direction reaches it (permutation $p{=}0.02$).}
\label{fig:specificity}
\end{figure}

\begin{figure}[t]
\centering
\includegraphics[width=\columnwidth]{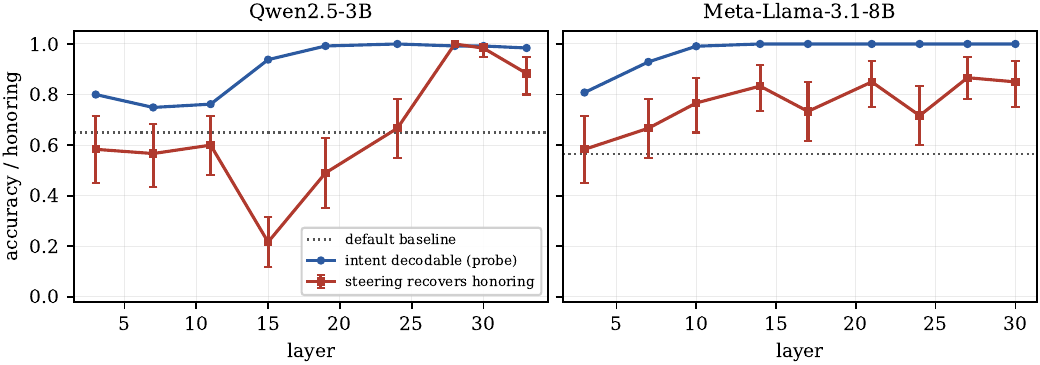}
\caption{\textbf{Represented before routed.} Per layer, how decodable the intent is (blue, probe accuracy) and
how much steering \emph{at that layer} recovers recognize-intent honoring (red, with bootstrap $95\%$ CIs),
against the default baseline (dotted). On both models the probe saturates at layers where steering has not yet
lifted honoring above baseline; recovery arrives only deeper (late and sharp for Qwen-3B, earlier and more
diffuse for Llama-8B). The gap between the blue and red onsets is the discard. $n{=}60$.}
\label{fig:localize}
\end{figure}

\section{Ceiling-model geometry (inconclusive)}
\label{app:geometry}
We asked whether the across-model stratification (Section~\ref{sec:scale}) has a geometric signature: do
the ceiling models, which honor recognize-intent by default, align their intent representation with the readout
more than the discard models do? We pre-registered, before running, the metric $M = |\cos(\text{intent
direction}, \text{readout direction})|$, where the intent direction is the final-layer logistic
recognize-vs-evaluate weight and the readout direction is the fixed acknowledgment-minus-feedback opener
unembedding difference, with the decision rule that the three ceiling models' $M$ must lie clearly above the
three discard models'. \emph{The result does not meet the rule.} Discard models: $M \in \{0.12, 0.06, 0.07\}$;
ceiling models: $\{0.14, 0.30, 0.00\}$. Two of three ceiling models (Mistral $0.30$, Qwen-14B $0.14$) exceed
the discard range, but Phi-3.5 is the lowest of all six ($0.00$), so there is no clean separation; the group
means differ ($0.15$ vs $0.08$) but the pre-registered separation criterion fails. A secondary operationalization
(the direction separating honored from discarded default replies, computable only where both classes occur)
gives $M \le 0.08$ throughout with no pattern. Per our pre-commitment we report the geometry as inconclusive and
make no mechanism claim from it; the across-model claim rests on the behavioral stratification and the
depth localization alone.

\section{Opener-biasing control (full numbers)}
\label{app:opener}
For the deflationary control of Section~\ref{sec:spec}, at each discard model's validated steer layer, matched
activation norm, generating at a $100$-token budget (longer than the $45$-token specificity run so the reply has
a body), we compare four quantities over the same $24$-item subset: $S_{\mathrm{true}}$, the behavior separation
from steering the learned intent direction; $S_{\mathrm{opener}}$, from steering the
acknowledgment-minus-feedback opener-unembedding direction directly; $S_{\mathrm{rest}}$, $S_{\mathrm{true}}$
recomputed with each reply's first sentence removed; and $|\cos|$ between the intent and opener directions at
that layer.

\begin{center}\small
\begin{tabular}{lccccc}
\toprule
model & layer & $S_{\mathrm{true}}$ & $S_{\mathrm{opener}}$ (coherent) & $S_{\mathrm{rest}}$ & $|\cos|$ \\
\midrule
Qwen2.5-7B   & 16 & 14 & $-16$ (only $2/24$ coherent) & 14 & 0.001 \\
Llama-3.1-8B & 19 &  7 & $1$ ($24/24$ coherent, inert) &  4 & 0.034 \\
Qwen2.5-3B   & 30 &  4 & $-1$ (incoherent)            & $-2$ & 0.138 \\
\bottomrule
\end{tabular}
\end{center}

On Qwen-7B the opener direction at matched norm degrades coherence before it honors ($2/24$ coherent), and
$S_{\mathrm{rest}}$ equals the full $S_{\mathrm{true}}$, so intent steering reorients the body, not the opener.
On Llama-8B the longer budget lifts the true separation clear of the random ceiling ($S{=}7$ vs $\max|S|{=}3$)
where the $45$-token run left it tied, the opener direction is inert ($S_{\mathrm{opener}}{=}1$), and
$S_{\mathrm{rest}}{=}4$ shows a modest body effect. On Qwen-3B the true separation \emph{dilutes} at this budget
($S{=}4$, below $\max|S|{=}12$): steered toward acknowledgment the small model acknowledges first but drifts
back into feedback over a long reply (negative-side feedback rises from $3/24$ at $45$ tokens to $17/24$ at
$100$), so the body test is inconclusive on 3B and its refutation rests on the norm-independent
near-orthogonality ($|\cos|{=}0.14$) and the opener direction's incoherence. At the $45$-token budget of
Section~\ref{sec:spec} the 3B separation is strong ($S_{\mathrm{true}}{=}14$), so the dilution is a
generation-length effect, and the intervention on the smallest model is front-loaded. The load-bearing evidence
against opener-biasing is the near-zero cosine on all three models and the body reorientation on Qwen-7B (and,
more modestly, Llama-8B). Script: \texttt{experiments/modal\_opener\_control.py}.

\section{Cross-model direction transport (exploratory)}
\label{app:transport}
The models do not share a hidden size, so a direction cannot transfer verbatim. We fit a linear map from
Qwen-3B activations to Llama-8B activations on a train split of stimuli and transport Qwen's intent direction
into Llama's space. The transported direction lands at cosine $0.85$ to Llama's \emph{own} intent direction,
and steering Llama with it recovers honoring ($0.56 \to 1.00$ on held-out items). We keep this out of the main
text and flag it as exploratory: the map's held-out activation $R^2$ is negative (it aligns the intent
direction, it does not reconstruct activation space), and a random direction of matched norm gives partial
non-specific recovery ($0.69$), so the signal is the direction alignment, not the raw steering number. Read as
suggestive that the intent axis is shared up to a linear map, not as an isomorphism claim.

\section{Request-matched construct control (full numbers)}
\label{app:request}
For the construct-validity control of Section~\ref{sec:repr}, both intent classes carry an explicit directive
so that request-presence is held constant and only the content of the request differs; recognize prefixes
become requests for acknowledgment (``please just celebrate this with me, I really don't want any notes''; ``do
me a favor and just take it in, don't give me feedback''; eight in all), paired with the unchanged evaluate
directives and the same surface-matched suffix. Probe accuracy is leave-one-phrasing-out CV at three depths;
the bag-of-words baseline uses the same phrasing folds.

\begin{center}\small
\begin{tabular}{lcccc}
\toprule
model & probe ($0.5$ / $0.67$ / $0.83$ depth) & peak & bag-of-words \\
\midrule
Qwen2.5-3B   & 0.912 / 0.898 / 0.973 & 0.973 & 0.688 \\
Qwen2.5-7B   & 0.929 / 0.891 / 0.955 & 0.955 & 0.688 \\
Llama-3.1-8B & 0.950 / 0.968 / 0.943 & 0.968 & 0.688 \\
Phi-3.5-mini & 1.000 / 0.984 / 0.953 & 1.000 & 0.688 \\
\bottomrule
\end{tabular}
\end{center}

The probe clears the bag-of-words baseline by roughly $0.25$--$0.30$ on every model with request-presence
matched. The baseline is higher than the surface-matched set's $0.48$ because celebration-requests and
critique-requests use different content words; that residual lexical signal is exactly what the surface-matched
design (Table~\ref{tab:probe}) controls, and the two designs together rule out both confounds. Script:
\texttt{experiments/modal\_request\_matched.py}.

\section{Second intent axis: vent vs solve (full numbers)}
\label{app:axis2}
To test that the represents-discard-recover chain is not specific to the recognize-versus-evaluate contrast, we
replicate it on a different intent: \emph{vent} (``I'm not looking for advice, I just need to be heard'') versus
\emph{solve} (``give me concrete steps to fix this''), behavior scored by whether the reply offers a solution,
surface-matched stimuli built as before on Qwen2.5-3B and Llama-3.1-8B. Intent decodes at probe $1.00$ (Qwen-3B)
and $0.97$ (Llama-8B) at the surface-matched token. At $n{=}60$ with bootstrap CIs, Qwen-3B honors the vent
intent only $0.70$ by default and steering recovers it to $0.98$ (non-overlapping); Llama-8B already honors it
at $0.88$ baseline (ceiling), so steering only nudges it to $0.98$. The representation holds on both; the clean
discard-then-recovery is on the model that discards. This axis is more lexically marked than the first: a
bag-of-words classifier reaches $0.79$ on the full text (versus $0.46$--$0.48$ on recognize-vs-evaluate), so it does
more to show the behavior generalizes to a second intent than to re-establish surface-independent
representation.

\begin{center}\small
\begin{tabular}{lcccc}
\toprule
model & probe & vent honored (default) [95\% CI] & steered [95\% CI] & default vs steered \\
\midrule
Qwen2.5-3B   & 1.00 & 0.70 [0.58, 0.82] & 0.98 [0.95, 1.00] & non-overlapping \\
Llama-3.1-8B & 0.97 & 0.88 [0.80, 0.95] & 0.98 [0.95, 1.00] & ceiling \\
\bottomrule
\end{tabular}
\end{center}

\begin{table}[h]
\centering\small
\begin{tabular}{lc}
\toprule
condition & recognize honored [95\% CI] \\
\midrule
default (no prompt) & 0.65 [0.53, 0.77] \\
mild nudge & 0.45 [0.33, 0.58] \\
explicit intent prompt & 0.93 [0.87, 0.98] \\
steer (no prompt) & 0.98 [0.95, 1.00] \\
explicit prompt $+$ steer & \textbf{1.00} [1.00, 1.00] \\
\bottomrule
\end{tabular}
\caption{Prompting vs routing on Qwen2.5-3B, $n{=}60$, bootstrap $95\%$ CI (full numbers for
Section~\ref{sec:prompt}). An explicit prompt ($0.93$) and steering ($0.98$) are interchangeable (overlapping
intervals) and stack ($1.00$): the represented direction is a handle as good as the instruction, with no
prompt. A vague nudge backfires ($0.65 \to 0.45$).}
\label{tab:prompt}
\end{table}

\section{Third intent axis: support vs help (full numbers)}
\label{app:axis3}
The support-versus-help axis (Section~\ref{sec:axis3}) is inferred and lexically clean. Per model:

\begin{center}\small
\begin{tabular}{lcccc}
\toprule
 & \multicolumn{2}{c}{represents} & \multicolumn{2}{c}{solutions offered} \\
\cmidrule(lr){2-3}\cmidrule(lr){4-5}
model & probe & BoW & on support & on help \\
\midrule
Qwen2.5-3B   & 0.74 & 0.57 & 0.32 & 0.50 \\
Qwen2.5-7B   & 0.71 & 0.57 & 0.28 & 0.40 \\
Qwen2.5-14B  & 0.86 & 0.57 & 0.32 & 0.42 \\
Mistral-7B   & 0.80 & 0.57 & 0.22 & 0.47 \\
Phi-3.5-mini & 0.82 & 0.57 & 0.47 & 0.57 \\
Llama-3.1-8B & 0.82 & 0.57 & 0.10 & 0.15 \\
\bottomrule
\end{tabular}
\end{center}
\noindent Bag-of-words sits at $0.57$ against probe $0.71$--$0.86$ on all six models, so the \emph{representation}
generalizes to a genuinely different, fully inferred intent. Behaviorally the readout tracks the intent only
weakly (help-intent draws unsolicited solutions more than support-intent), and steering the direction
\emph{fails} the specificity control (Section~\ref{sec:axis3}), so we claim the representation-generalization
here, not a causal handle. $n{=}40$ per intent.

\section{Per-model probe, bag-of-words, and steer layer}
\label{app:general}
Table~\ref{tab:general} tabulates the per-model probe, bag-of-words, and steer-layer figures referenced in
Section~\ref{sec:scale}, across all six models and four families.

\begin{table}[h]
\centering\small
\begin{tabular}{lccc}
\toprule
model & layers & probe / BoW & steer (depth) \\
\midrule
Qwen2.5-3B  & 36 & 1.00 / 0.48 & 30 (0.83) \\
Qwen2.5-7B  & 28 & 1.00 / 0.46 & 16 (0.57) \\
Qwen2.5-14B & 48 & 1.00 / 0.46 & 41 (0.85) \\
Mistral-7B  & 32 & 1.00 / 0.46 & 22 (0.69) \\
Phi-3.5-mini & 32 & 1.00 / 0.46 & 22 (0.69) \\
Llama-3.1-8B & 32 & 1.00 / 0.46 & 19 (0.59) \\
\bottomrule
\end{tabular}
\caption{Representation and steer layer across six models, four families (recognize/evaluate axis). Intent
decodes at probe $1.00$ with bag-of-words at chance ($\le 0.48$) everywhere: the representation is not a
small-model or single-family artifact. The steer layer, and its fraction of depth, is model-specific. The
behavioral discard and recovery, model-specific, are quantified at $n{=}60$ in Table~\ref{tab:scale}. A seventh
model, Qwen2.5-32B, is run probe-and-honoring only: it represents the intent (probe $0.99$ stated, $0.93$
inferred) and honors it at baseline ($0.78$/$0.77$), so it has no discard to steer (Section~\ref{sec:scale}).}
\label{tab:general}
\end{table}

\section{Naturalistic transfer (full numbers)}
\label{app:natural}
A reader may suspect the synthetic, surface-matched stimuli drive the effect. We test whether the probe and the
steering direction, both fit only on templates, transfer to hand-written non-templated messages (varied length,
register, and topic; e.g.\ ``just got back from my first 5k, didn't stop once''). On $23$ such messages the
template-trained probe classifies their intent at $1.00$ (layer~24), and the behavioral chain reproduces:
default honoring $0.62$ (the same discard), steering recovers it to $1.00$ ($23/23$ coherent), and
genuinely-evaluate messages correctly draw feedback ($0.08$ honoring). A direction learned on templates governs
behavior on real messages; the sample is small ($n{=}23$, hand-written by the author).

\section{Reproducibility}
\label{app:repro}
\begin{tcolorbox}[graybox, breakable, title={Reproduce}]
\begin{verbatim}
pip install -e .
# core chain (CPU; downloads Qwen2.5-3B-Instruct)
python experiments/probe_intent.py        # represents (leave-phrasing-out)
python experiments/same_model_discard.py  # discards (same model)
python experiments/steer_probe_sweep.py   # find the causal layer
python experiments/steer_dose.py          # dose-response confirmation
pytest                                     # pipeline + no-leak guards
# extended experiments (GPU, via Modal)
modal run experiments/modal_sweep.py      # generality ladder (6 models)
modal run experiments/modal_scale.py      # discard/recover, n=60
modal run experiments/modal_specificity2.py  # spec controls
modal run experiments/modal_localize2.py  # depth localization
modal run experiments/modal_natural.py    # naturalistic transfer
\end{verbatim}
\end{tcolorbox}
The core probe is CPU-only and downloads Qwen2.5-3B-Instruct; the extended experiments run on a single
A100 through Modal (\url{https://modal.com}), and the behavioral elicitation uses API keys. Code, stimuli, and
all experiment scripts are available at
\ifsubmission \url{https://anonymous.4open.science/r/recipient-probe}\else
\url{https://github.com/collapseindex/recipient-probe}\fi.

\paragraph{Software.} Experiments use PyTorch \citep{paszke2019pytorch}, HuggingFace Transformers
\citep{wolf2020transformers}, scikit-learn \citep{pedregosa2011sklearn} for the probes and steering fits, and
sentence-transformers \citep{reimers2019sbert} for the independent embedding measure (Section~\ref{sec:discard}).

\end{document}